\pdfoutput=1

\documentclass[11pt, a4paper]{article}

\usepackage[margin=1in]{geometry}

\usepackage{times}
\usepackage{latexsym}
\usepackage[round]{natbib}
\usepackage{xcolor}
\usepackage[
    colorlinks=true,
    citecolor=blue!70!black,
    linkcolor=blue!70!black,
    urlcolor=blue!70!black
]{hyperref}

\usepackage{url}

\usepackage[T1]{fontenc}
\usepackage[utf8]{inputenc}
\usepackage{textcomp}

\usepackage{microtype}

\usepackage{inconsolata}

\usepackage{graphicx}
\usepackage{booktabs}
\usepackage{multirow}
\usepackage{amsmath}

\usepackage{setspace}
\title{\textbf{NCTB-QA: A Large-Scale Bangla Educational Question Answering Dataset and Benchmarking Performance}}

\author{
Abrar Eyasir\textsuperscript{+} \quad Tahsin Ahmed\textsuperscript{+} \quad Muhammad Ibrahim\thanks{Corresponding author: ibrahim313@du.ac.bd} \\
\vspace{0.3em}
\textit{Department of Computer Science and Engineering} \\
\textit{University of Dhaka} \\
\vspace{0.3em}
\texttt{\{eyasir2047, tahsin030602\}@gmail.com, ibrahim313@du.ac.bd}
}

\usepackage{fancyhdr}
\date{}

\begin{document}
\maketitle
\begin{abstract}
Reading comprehension systems for low-resource languages face significant challenges in handling unanswerable questions. These systems tend to produce unreliable responses when correct answers are absent from context. To solve this problem, we introduce NCTB-QA, a large-scale Bangla question answering dataset comprising 87,805 question-answer pairs extracted from 50 textbooks published by Bangladesh's National Curriculum and Textbook Board. Unlike existing Bangla datasets, NCTB-QA maintains a balanced distribution of answerable (57.25\%) and unanswerable (42.75\%) questions. NCTB-QA also includes adversarially designed instances containing plausible distractors. We benchmark three transformer-based models (BERT, RoBERTa, ELECTRA) and demonstrate substantial improvements through fine-tuning. BERT achieves 313\% relative improvement in F1 score (0.150 to 0.620). Semantic answer quality measured by BERTScore also increases significantly across all models. Our results establish NCTB-QA as a challenging benchmark for Bangla educational question answering. This study demonstrates that domain-specific fine-tuning is critical for robust performance in low-resource settings.
\end{abstract}

\section{Introduction}

Question answering (QA) systems have made strong progress in high-resource languages, largely due to the availability of large datasets such as SQuAD \citep{rajpurkar-etal-2016-squad}, SQuAD 2.0 \citep{rajpurkar-etal-2018-know}, HotpotQA \citep{yang-etal-2018-hotpotqa}, and Natural Questions \citep{kwiatkowski-etal-2019-natural}. In contrast, low-resource languages remain underexplored, even though they are spoken by a large share of the world’s population. Bangla, the seventh most spoken language with over 230 million speakers, is a clear example. Existing Bangla QA datasets are limited in size and coverage. These datasets show little emphasis on educational content or unanswerable questions.

For real-world use, QA systems must also know when a question cannot be answered from the given context.  \citet{rajpurkar-etal-2018-know} demonstrates that models trained only on answerable questions often respond with high confidence even when the required information is missing. This hallucination behaviour of models results in producing fluent but incorrect answers. This issue is quite serious in educational settings. Incorrect responses can confuse learners and reduce trust in automated systems.

Most existing Bangla QA datasets are small in scale. They typically contain between 3,000 \citep{khondoker2024unlockingpotentialmultiplebert} and 14,889 \citep{ekram-etal-2022-banglarqa} question–answer pairs, which is not sufficient for training robust transformer-based models. Second, although TigerLLM \citep{raihan-zampieri-2025-tigerllm} is considerably larger, containing nearly 10 million tokens collected from 163 textbooks for grades 6–12, it does not include unanswerable questions. This limits its usefulness for evaluating a model’s ability to abstain from hallucinated responses. Third, BanglaRQA \citep{ekram-etal-2022-banglarqa} includes only a small number of factoid-style unanswerable questions and lacks adversarially constructed cases. More broadly, many existing Bangla QA datasets focus predominantly on factoid questions. These questions restrict linguistic diversity and limit deeper reasoning-based evaluation.

To address these limitations, we introduce NCTB-QA, a large-scale Bangla question answering dataset designed specifically for educational use. The dataset contains 87,805 question–answer pairs collected from 50 textbooks published by Bangladesh’s National Curriculum and Textbook Board (NCTB), covering grades 1 through 10. NCTB-QA maintains a balanced distribution of answerable and unanswerable questions. NCTB-QA also introduces unanswerable instances adversarially designed to include plausible distractors. The adversarial questions force the model to understand genuine contextual reasoning. By sourcing all content directly from official NCTB materials, the dataset remains closely aligned with real educational curricula.

\section{Related Work}

\subsection{English Question Answering}

The Stanford Question Answering Dataset (SQuAD) introduces more than 100,000 crowd-written questions based on Wikipedia articles \citep{rajpurkar-etal-2016-squad} and establishes extractive reading comprehension as a central benchmark for NLP research. However, SQuAD contains only answerable questions, which often lead models to produce plausible but fabricated answers when no correct response is present in the context.

To address this limitation, \citet{rajpurkar-etal-2018-know} introduces unanswerable questions in the SQuAD~2.0 dataset. These additions force models to learn when to refrain from answering. The dataset comprises 130,319 examples, including 43,498 unanswerable questions. This results in a more balanced evaluation setting.

\subsection{Bangla Question Answering}

Bangla QA research has improved gradually; however, available datasets remain small and limited in scope. An early work, Multiple BERT Models for Bangla Questions in NCTB Textbooks, introduces approximately 3,000 human-annotated question–answer pairs \citep{khondoker2024unlockingpotentialmultiplebert}. However, it includes only answerable questions and remains limited in size.

BanglaRQA represents a significant step forward for Bangla question answering. It contains 3,000 passages and 14,889 QA pairs, including both answerable and unanswerable questions \citep{ekram-etal-2022-banglarqa}. Among these, 11,258 are answerable and 3,631 are unanswerable. Transformer-based models such as mBERT, BanglaBERT, mT5, and BanglaT5 are evaluated, achieving a best performance of 62.42\% Exact Match and 78.11\% F1. However, the imbalance between answerable and unanswerable questions (approximately 3:1) limits the dataset’s effectiveness for robust answerability detection.

TigerLLM focuses on language modeling using 10 million tokens from NCTB textbooks but does not include unanswerable questions \citep{raihan-zampieri-2025-tigerllm}. More recently, the BEnQA benchmark introduces approximately 5,161 Bangla–English exam questions from school science subjects \citep{shafayat-etal-2024-benqa}. It includes both factual and reasoning questions. This study finds that models perform worse in Bangla than in English.

Overall, existing Bangla QA datasets lack sufficient scale, domain focus, and balanced answerable and unanswerable examples, highlighting a clear gap in Bangla QA resources. Table~\ref{tab:dataset_comparison} compares NCTB-QA with previous datasets.

\begin{table*}[t]
\centering
\small
\begin{tabular}{lcccccc}
\toprule
\textbf{Dataset} & \textbf{QA Pairs} & \textbf{Contexts} & \textbf{Answerable} & \textbf{Unanswerable} & \textbf{Domain} \\
\midrule
\textbf{NCTB-QA (Ours)} & 87,805 & 10,070  & 50,270 (57.25\%) & 37,535 (42.75\%) & Educational \\
\midrule
TigerLLM (\citeauthor{raihan-zampieri-2025-tigerllm}) & 10M tokens & 163 textbooks & 100\% & 0\% & Educational \\
\midrule
BanglaRQA (\citeauthor{ekram-etal-2022-banglarqa}) & 14,889 & 3,000  & 11,258 (75.6\%) & 3,631 (24.4\%) & General \\
\midrule
BEnQA (\citeauthor{shafayat-etal-2024-benqa}) & 5,161 & -- & 100\% & 0\% & Educational \\
\midrule
NCTB(\citeauthor{khondoker2024unlockingpotentialmultiplebert}) & 3,000 & -- & 3,000 (100\%) & 0\% & Educational \\
\midrule
BanglaQA (\citeauthor{shahriar-etal-2023-question}) & 178,012 & 15,997 articles & 142,181 (79.87\%) & 35,831 (20.13\%) & News \\
\bottomrule
\end{tabular}
\caption{Comparison of NCTB-QA with existing Bangla question–answering datasets. NCTB-QA uniquely combines large-scale, educational domain specificity, and a relatively balanced distribution of answerable and unanswerable questions.}
\label{tab:dataset_comparison}
\end{table*}

\section{NCTB-QA}

\subsection{Dataset Construction}

Textbooks for grades 1--10 were collected from the official website of the National Curriculum and Textbook Board (NCTB)\footnote{\url{https://nctb.gov.bd/site/page/934d6561-1fb4-466c-a0e9-330b748098e6}}. To ensure reproducibility and consistency, the collection process was fully automated using a web-scraping pipeline rather than manual downloading.

We iterated through class-wise textbook listings and extracted subject-specific book entries. Downloaded files were stored in a hierarchical directory structure organized by class. We enforced short delays between requests to avoid server overload.

Duplicate textbooks were automatically detected and removed. This ensures that each textbook appeared exactly once in the corpus. The final dataset consists of 50 unique textbooks. These textbooks cover a wide range of academic subjects, including language arts, science, and social studies. They serve as the primary source material for downstream content extraction and QA generation.

\subsection{Content Extraction}

\begin{figure*}[t]
\centering
\includegraphics[width=0.85\linewidth]{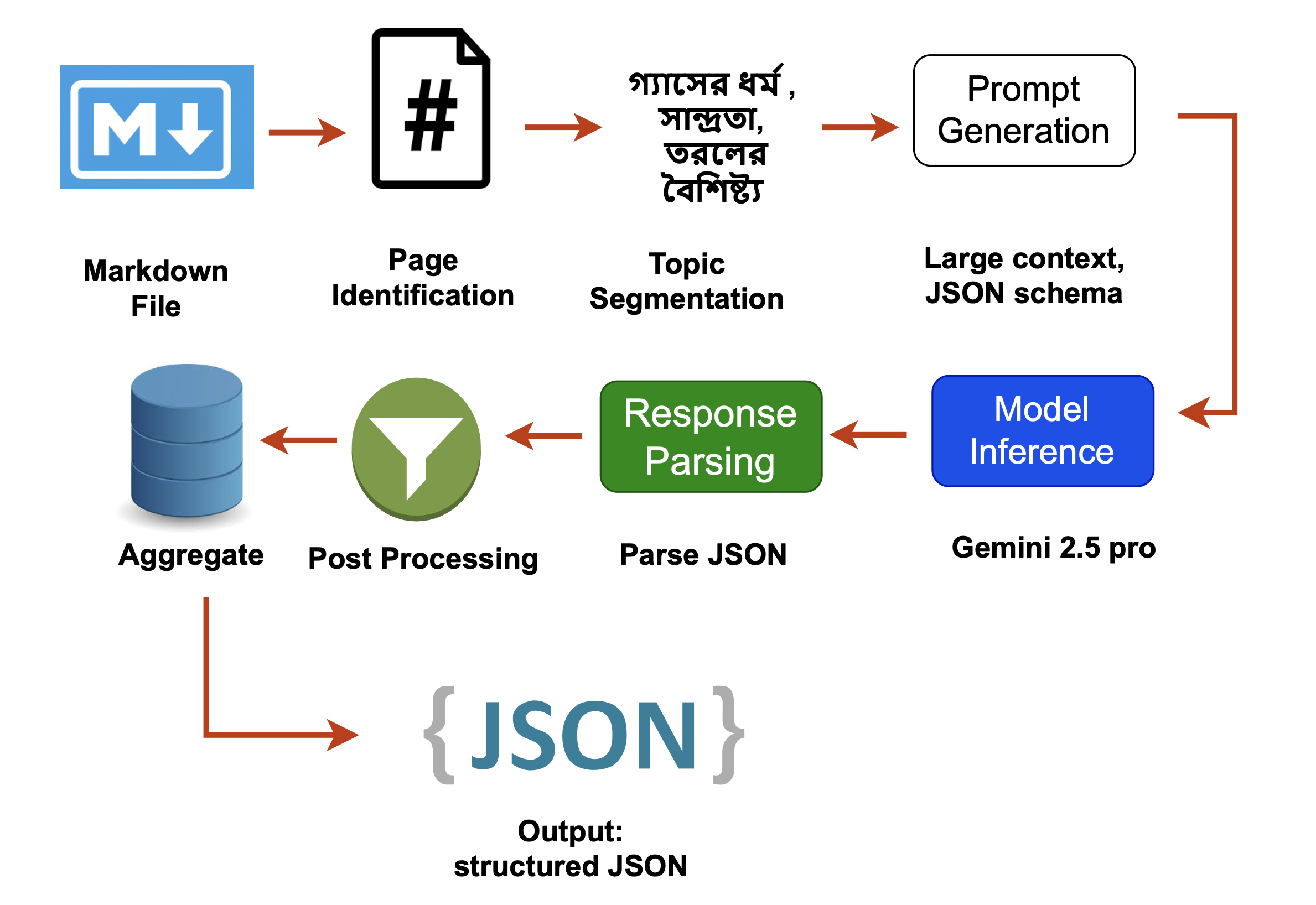}
\caption{Overview of the content extraction and QA generation pipeline. Markdown textbooks are cleaned, segmented, and thematically grouped into coherent contexts, which are then processed by Gemini to generate structured QA pairs in JSON format.}
\label{fig:qa_process}
\end{figure*}

All textbooks were processed in their text-based Markdown format. Before QA generation, we applied a multi-stage content extraction and cleaning pipeline to transform raw Markdown files into high-quality contextual segments suitable for language model inference, as illustrated in Figure~\ref{fig:qa_process}.

First, raw Markdown files were cleaned to remove non-instructional artifacts such as page numbers, formatting noise, extraction errors, exercise questions, and metadata. This cleaning step preserved the core explanatory and narrative educational content while discarding elements not useful for reading comprehension tasks.

Next, the cleaned Markdown documents were segmented into coherent textual contexts. We implemented a line-aware segmentation strategy with a maximum chunk size of 2000 characters and an overlap of 500 characters between adjacent segments to preserve contextual continuity. Segmentation respected paragraph and line boundaries to prevent semantic fragmentation.

To further improve semantic coherence, each chunk was analyzed using a language model to identify logical topic boundaries. Content was divided into thematically consistent sections based on conceptual similarity. We applied additional quality filtering by removing segments shorter than 150 characters.

The final output of this stage is a collection of high-quality, semantically coherent contextual segments stored in structured JSON format. They serve as direct input for QA generation.

\subsection{Question--Answer Generation}
We generated a large-scale Bangla QA dataset from the extracted contextual segments using Gemini~2.5~Pro. For each context, the model was prompted to generate diverse question types. The questions are semantically aligned with the context.

The generation process produced between 2 and 15 questions per context, resulting in an average of 8.74 questions. We enforced a near-equal balance between answerable and unanswerable questions, achieving a ratio of  1.34 :1 (50,270  answerable vs.\ 37,535 unanswerable).

For answerable questions, answers were constrained to be extractable text spans within the context, with precise character-level start and end offsets. Unanswerable questions were carefully constructed to remain topically relevant while ensuring that the context contained information of the appropriate answer type but not the correct answer itself. We explicitly discouraged trivial cues such as low lexical overlap, mismatched answer types, or unrelated topics.

Quality control included three automated validation steps: (i) minimum context length enforcement, (ii) verification that answer spans occurred at the specified offsets, and (iii) duplicate question detection. All QA pairs were serialized in the SQuAD-style JSON format~\citep{rajpurkar-etal-2016-squad}.

In total, the pipeline processed 11,344 contextual segments and generated 87,805 QA pairs. The resulting dataset constitutes a high-quality Bangla educational QA corpus suitable for reading comprehension and QA research.

\section{Dataset Analysis}
\label{sec:dataset-analysis}

We present a comprehensive analysis of the NCTB-QA dataset. We examine its distributional properties, question characteristics, and subject-matter coverage. This analysis provides insights into the complexity and diversity of NCTB-QA. This establishes its suitability for evaluating Bangla reading comprehension systems.

\subsection{Overall Statistics}

\begin{figure*}[t]
\centering
\includegraphics[width=0.8\linewidth]{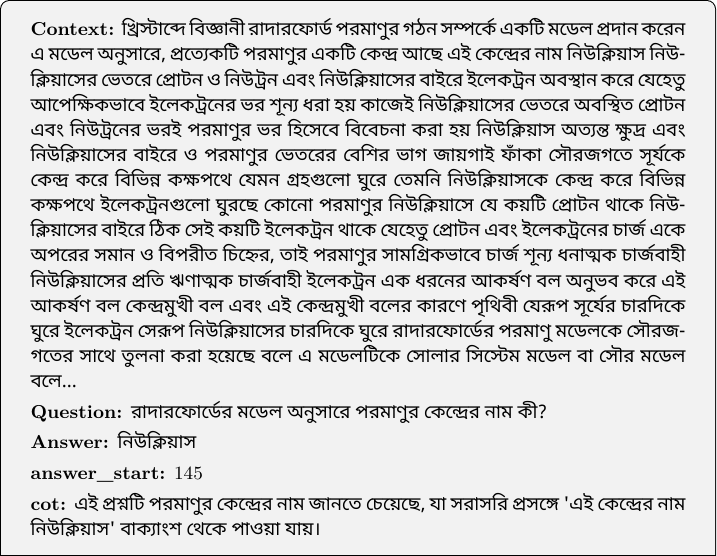}
\caption{Sample of Dataset}
\label{fig:sample_dataset}
\end{figure*}

Table~\ref{tab:overall_stats} presents the key statistics of NCTB-QA. The dataset comprises 87,805 question-answer pairs derived from 10,070 contextual segments across 50 source textbooks spanning grades 1--10. A notable characteristic is the balanced distribution between answerable (57.25\%) and unanswerable (42.75\%) questions. This ensures that models must learn both answer extraction and answerability detection capabilities.
Figure~\ref{fig:sample_dataset} shows an example from our dataset.

\begin{table}[t]
\centering
\small
\begin{tabular}{lr}
\toprule
\textbf{Metric} & \textbf{Value} \\
\midrule
Total QA pairs & 87,805 \\
Answerable questions & 50,270 (57.25\%) \\
Unanswerable questions & 37,535 (42.75\%) \\
Contextual segments & 10,070  \\
Source textbooks & 50 \\
Grade levels & 1--10 \\
\bottomrule
\end{tabular}
\caption{Overall statistics of the NCTB-QA dataset.}
\label{tab:overall_stats}
\end{table}

\subsection{Context and Question Characteristics}

\begin{figure}[t]
\centering
\includegraphics[width=0.8\linewidth]{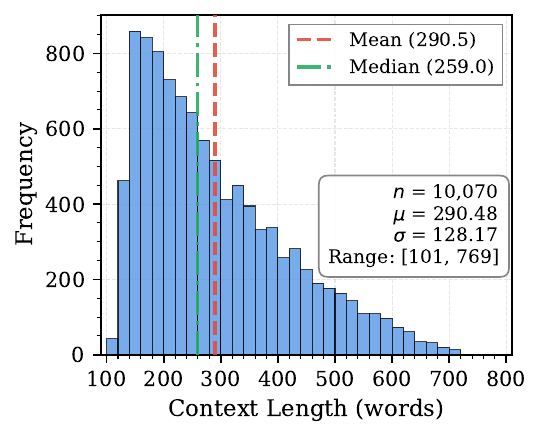}
\caption{Distribution of word count per context in NCTB-QA. The distribution spans from 103 to 1,871 words, with the majority of contexts containing between 150--400 words.}
\label{fig:context_length}
\end{figure}

\paragraph{Context Length Distribution.} Figure~\ref{fig:context_length} presents the distribution of context lengths in NCTB-QA. The statistics show a well-controlled and moderately right-skewed distribution. The mean context length is 290.48 words, while the median is 259 words, indicating that most contexts are concentrated around 250--300 words, with a smaller proportion of longer contexts forming the right tail. The standard deviation is 128.17 words. This reflects moderate variability without extreme dispersion. Context lengths range from a minimum of 101 words to a maximum of 769 words. This demonstrates the absence of both trivially short and excessively long contexts. Percentile analysis further supports this observation. Specifically, 50\% of contexts fall between 188 and 369 words (interquartile range). Again, 90\%, 95\%, and 99\% of contexts contain fewer than 482, 547.55, and 644 words, respectively. This confirms that longer contexts are relatively rare and do not dominate the dataset

\begin{figure}[t]
\centering
\includegraphics[width=0.8\linewidth]{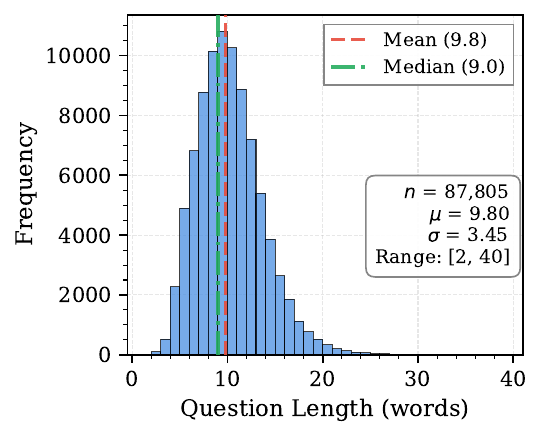}
\caption{Distribution of word count per question in NCTB-QA. Questions are predominantly concise, with most containing between 7--12 words.}
\label{fig:question_length}
\end{figure}

\paragraph{Question Length Distribution.} As shown in Figure~\ref{fig:question_length}, questions in NCTB-QA are relatively concise, with a mean length of 9.80 words (SD = 3.45) and a median of 9 words. The distribution is approximately normal, with 75\% of questions containing 12 or fewer words. The peak frequency occurs at 9 words (12.32\% of questions), which demonstrates consistency in question formulation. This conciseness aligns with natural educational question-asking patterns while maintaining sufficient complexity for comprehension assessment.

\begin{figure}[t]
\centering
\includegraphics[width=0.8\linewidth]{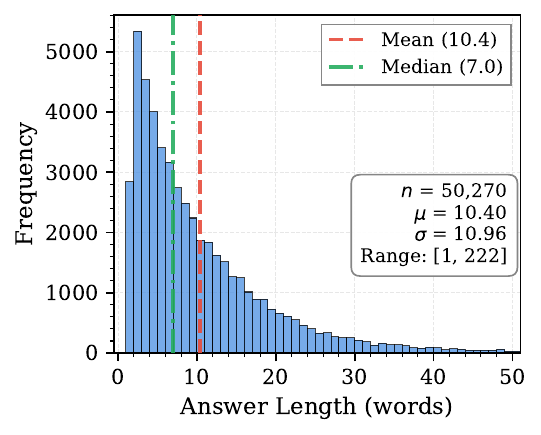}
\caption{Distribution of word count per answer in NCTB-QA. The right-skewed distribution shows most answers are brief, while some require extended responses.}
\label{fig:answer_length}
\end{figure}

\paragraph{Answer Length Distribution.} Figure~\ref{fig:answer_length} presents the distribution of answer lengths for the 50,270 answerable questions. Answers are more variable than questions, with an average length of 10.40 words and a median of 7 words. The distribution is right-skewed: most answers are short (75\% contain 14 words or fewer), while a small number extend to over 200 words. Notably, 30.29\% of answers are 1--3 words long. This indicates a substantial proportion of factual questions require brief responses. Again, the long tail reflects questions demanding more elaborate explanations.

\begin{figure}[t]
\centering
\includegraphics[width=0.8\linewidth]{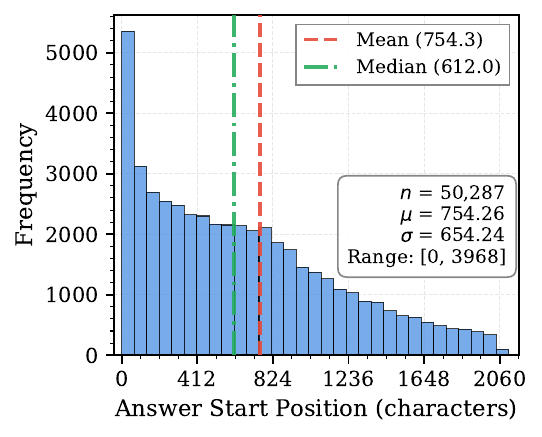}
\caption{Distribution of answer start positions (character index) in NCTB-QA. Answers are distributed throughout contexts, mitigating positional bias. }
\label{fig:answer_position}
\end{figure}

\paragraph{Answer Position Distribution.} Figure~\ref{fig:answer_position} shows the distribution of answer start positions within contexts (measured in character indices). The mean position is 754.26 characters (SD = 654.24), with a median of 612 characters. While answers are distributed throughout the contexts, there is a slight tendency toward earlier positions, with 25\% of answers beginning within the first 240 characters. However, 25\% of answers start beyond character 1,079, and some answers appear as late as character 3,968. This demonstrates that the dataset requires models to search comprehensively throughout the entire context rather than relying on positional biases.

\subsection{Subject Distribution}

\begin{figure*}[t]
\centering
\includegraphics[width=\textwidth]{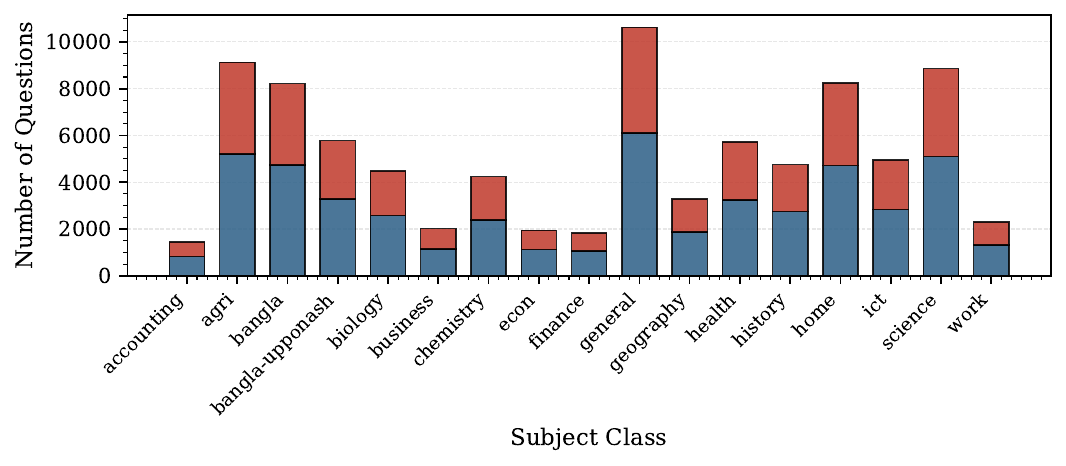}
\caption{Subject-wise distribution of answerable (blue) and unanswerable (red) questions in NCTB-QA, showing consistent answerability ratios across domains.}
\label{fig:subject_distribution}
\end{figure*}

Figure~\ref{fig:subject_distribution} reports the distribution of NCTB-QA across 17 curricular subjects, decomposed into answerable and unanswerable instances. The dataset exhibits broad topical coverage with consistent supervision balance across domains. General Studies comprises the largest share with 10,621 examples, of which 6,102 are answerable and 4,519 are unanswerable (57.45\% answerable). Agriculture and Home Science follow with 9,129 examples (5,213/3,916) and 8,245 examples (4,711/3,534), respectively. Bangla language and literature are strongly represented, including general Bangla with 8,216 instances (4,745/3,471) and Bangla novels with 5,784 instances (3,288/2,496).

STEM disciplines account for a substantial fraction of the corpus, including Science with 8,857 instances (5,093 answerable, 3,764 unanswerable), Biology with 4,490 (2,570/1,920), Chemistry with 4,242 (2,393/1,849), and ICT with 4,944 (2,832/2,112). Humanities and social sciences further broaden coverage, notably History with 4,752 examples (2,740/2,012), Geography with 3,291 (1,882/1,409), Economics with 1,939 (1,118/821), and Business Studies with 2,013 (1,154/859).

The proportion of answerable questions remains tightly concentrated around 57\% across all subjects (std. dev. < 0.6\%). This suggests controlled dataset construction and minimal subject-induced supervision skew. This uniformity, combined with wide curricular breadth, positions NCTB-QA as a domain-diverse benchmark for robust comprehension modeling in low-resource educational contexts.

\subsection{Question Type Analysis}
To better understand the contents of NCTB-QA,
we analyze the distribution of different question-answer types for train, validation, and test sets.
We partitioned NCTB-QA using an 80/10/10 split: 70,167 questions for training, 8,798 for validation, and 8,840 for testing. Splits maintained balanced answerable and unanswerable ratios with proportional representation across question types and subject areas. We ensured no context appeared in multiple splits to prevent data leakage.

Table~\ref{tab:question_types} categorizes questions by linguistic and semantic type across the train, validation, and test splits. We identify six primary question types: \textit{Factual} questions seeking specific information (23.65\% of all questions), \textit{Causal} questions exploring cause-effect relationships (7.07\%), \textit{Comparative} questions requiring comparison (5.00\%), \textit{Negation} questions involving negative constructions (10.66\%), \textit{Confirmation} questions seeking yes/no responses (0.13\%), and \textit{Others} capturing remaining question types (53.49\%).

The distribution remains consistent across splits. All splits maintain similar proportions of each question type. The ratio of unanswerable to answerable questions is relatively consistent within each type, though comparative questions show a higher proportion of unanswerable instances (54.46\% unanswerable vs. 45.54\% answerable). This suggests that comparative reasoning may be particularly sensitive to context coverage. The predominance of the "others" category (53.49\%) reflects the diversity of educational assessment questions beyond traditional taxonomies. The presence of factual (23.65\%), causal, comparative, and negation questions also ensures the dataset tests various reasoning capabilities.

\begin{table}[t]
\centering
\small
\begin{tabular}{lrr}
\toprule
\textbf{Split} & \textbf{Unanswerable} & \textbf{Answerable} \\
\midrule
\textbf{Train} \\
Factual        & 7,359 & 9,223 \\
Confirmation   & 15   & 67   \\
Causal         & 2,092 & 2,792 \\
Comparative    & 1,900 & 1,607 \\
Negation       & 3,441 & 4,116 \\
Others         & 15,218 & 22,473 \\
\midrule
\textbf{Validation} \\
Factual        & 899 & 1,124 \\
Confirmation   & 2   & 7    \\
Causal         & 300 & 356  \\
Comparative    & 232 & 208  \\
Negation       & 415 & 501  \\
Others         & 1,831 & 2,726 \\
\midrule
\textbf{Test} \\
Factual        & 959 & 1,206 \\
Confirmation   & 2   & 17   \\
Causal         & 296 & 373  \\
Comparative    & 259 & 184  \\
Negation       & 413 & 473  \\
Others         & 1,902 & 2,817 \\
\midrule
\textbf{Total} & \textbf{37,535} & \textbf{50,270} \\
\bottomrule
\end{tabular}
\caption{Distribution of question types across train, validation, and test splits in NCTB-QA. The consistent proportions across splits ensure balanced evaluation of different reasoning capabilities.}
\label{tab:question_types}
\end{table}

Overall, the dataset’s scale, subject coverage, and structural diversity make it suitable for evaluating Bangla reading comprehension systems across educational levels.

\section{Methodology}
\label{sec:methodology}

The task is reading comprehension-based QA. The model is given the question and its associated context passage as input and the model outputs the answer. If the question is unanswerable, then the output is an empty string. Three different models were implemented: BERT, RoBERTa, and ELECTRA. This section explains the whole pipeline of the experiments, from preprocessing the data to model training and evaluation. We report training hyperparameters and experimental setup for reproducibility.

\subsection{Data Preprocessing}
\label{subsec:preprocessing}

We applied a multi-stage preprocessing pipeline to prepare the SQuAD-format data for model training.

\textbf{Answer span correction.} We fixed misaligned \texttt{answer\_start} offsets by (1) verifying that the character span at \texttt{answer\_start} matches the answer text; (2) if not, searching for the answer via exact, case-insensitive, and whitespace-normalized matching; and (3) falling back to partial matching on the first three words when full matching fails. Entries with unfixable spans were discarded.

\textbf{Long-context windowing.} To handle contexts exceeding the model's maximum sequence length while preserving answer locality, we split each context into overlapping windows. We used a maximum of 2,000 tokens per window with a stride of 300 tokens. For each window, we retained only QA pairs whose answer spans fall entirely within that window. This converted absolute character offsets to window-relative offsets.

\textbf{Unanswerable handling.} Questions marked \texttt{is\_impossible} or with empty answer lists were treated as unanswerable and assigned empty answer spans. The pipeline supports SQuAD 2.0-style negative examples throughout.

\subsection{Model Architecture and Training Setup}
\label{subsec:models}

We fine-tuned three encoder-only transformer architectures for extractive QA: BERT-base-cased \citep{devlin-etal-2019-bert}, RoBERTa-base \citep{liu2019robertarobustlyoptimizedbert}, and ELECTRA-base-discriminator\citep{clark2020electrapretrainingtextencoders}. All models were implemented via the SimpleTransformers \texttt{QuestionAnsweringModel}. These models were trained to predict start and end token positions of answer spans within the concatenated question--context input.

\textbf{Training configuration.} We used memory-optimized settings suitable for 12GB VRAM: mixed-precision (FP16) training, no evaluation during training, and gradient accumulation to simulate larger batch sizes. Model-specific hyperparameters are listed in Table~\ref{tab:hyperparameters}. We used the AdamW optimizer with linear learning rate decay and warmup (10\% of steps), weight decay 0.01, and max gradient norm 1.0. For inference, we used \texttt{n\_best\_size}=3, \texttt{max\_answer\_length}=30, and \texttt{null\_score\_diff\_threshold}=0.0 to support unanswerable predictions.

\textbf{Implementation details.} Models were trained sequentially with full resource cleanup between runs to avoid out-of-memory errors. Inputs were tokenized with max sequence length of 384 and a document stride of 128. The pipeline was implemented in Python using PyTorch, Transformers, and SimpleTransformers.

\subsection{Evaluation Metrics}
\label{subsec:evaluation}

We reported three metrics on the test set.

\textbf{Exact Match (EM).} A prediction is correct if, after normalization (lowercasing, punctuation removal, whitespace collapse), it exactly matches any gold answer. For unanswerable questions, the model must predict an empty or unanswerable string to receive credit.

\textbf{F1 Score.} We computed token-level F1 between the predicted and gold answer, treating answers as multisets of tokens \citep{rajpurkar-etal-2016-squad}. Precision is the proportion of predicted tokens that appear in the gold answer. Again, recall is the proportion of gold tokens that appear in the prediction. F1 is the harmonic mean. Unanswerable questions were scored 1.0 if the model correctly abstained, else 0.0. 

\textbf{BERTScore.} We used BERTScore \citep{zhang2020bertscoreevaluatingtextgeneration} with the Bengali language code (\texttt{bn}) to measure semantic similarity between predictions and references. BERTScore F1 was computed in batches of 16 and averaged over answerable QA pairs. And unanswerable questions were handled separately (correct abstention = 1.0).

All metrics were computed with separate breakdowns for answerable and unanswerable subsets to support analysis of model behavior on both question types.

\subsection{Training Hyperparameters}
\label{subsec:hyperparameters}

Table~\ref{tab:hyperparameters} summarizes the training hyperparameters for each model. BERT and RoBERTa share similar settings; ELECTRA uses larger per-step batch sizes and fewer gradient accumulation steps due to its more memory-efficient architecture.

\begin{table*}[t]
\centering
\small
\begin{tabular}{lccc}
\toprule
\textbf{Hyperparameter} & \textbf{BERT} & \textbf{RoBERTa} & \textbf{ELECTRA} \\
\midrule
Pre-trained model & bert-base-cased & roberta-base & google/electra-base-discriminator \\
Max sequence length & 384 & 384 & 384 \\
Document stride & 128 & 128 & 128 \\
Train batch size & 2 & 2 & 4 \\
Eval batch size & 4 & 4 & 8 \\
Gradient accumulation steps & 16 & 16 & 8 \\
Effective batch size & 32 & 32 & 32 \\
Learning rate & $3 \times 10^{-5}$ & $2 \times 10^{-5}$ & $3 \times 10^{-5}$ \\
Epochs & 2 & 2 & 2 \\
Warmup ratio & 0.1 & 0.1 & 0.1 \\
Weight decay & 0.01 & 0.01 & 0.01 \\
Adam $\epsilon$ & $10^{-8}$ & $10^{-8}$ & $10^{-8}$ \\
Max gradient norm & 1.0 & 1.0 & 1.0 \\
Optimizer & AdamW & AdamW & AdamW \\
Scheduler & Linear + warmup & Linear + warmup & Linear + warmup \\
FP16 mixed precision & Yes & Yes & Yes \\
$n$-best size & 3 & 3 & 3 \\
Max answer length & 30 & 30 & 30 \\
\bottomrule
\end{tabular}
\caption{Training hyperparameters for BERT, RoBERTa, and ELECTRA extractive QA models.}
\label{tab:hyperparameters}
\end{table*}

\subsection{Experimental Setup}
\label{subsec:setup}

Experiments were conducted on a single GPU with 12GB VRAM. We validated the data files before training and built a ground-truth dictionary from the test set for evaluation purposes. Each model was evaluated before and after fine-tuning on the same preprocessed test data. Predictions were generated in batches of 16 to manage memory. Results were saved with per-model breakdowns (answerable vs.\ unanswerable EM/F1 and BERTScore) for analysis.

\section{Results}

\subsection{Overall Performance}

Table \ref{tab:main_results} presents performance before and after fine-tuning. Fine-tuning yields dramatic improvements across all metrics and models.

\begin{table*}[t]
\centering
\small
\begin{tabular}{llccc}
\toprule
\textbf{Model} & \textbf{Stage} & \textbf{EM} & \textbf{F1} & \textbf{BERTScore} \\
\midrule
\multirow{2}{*}{BERT} & Before & 0.059 & 0.150 & 0.665 \\
& After & \textbf{0.509} & \textbf{0.620} & \textbf{0.880} \\
\midrule
\multirow{2}{*}{RoBERTa} & Before & 0.427 & 0.428 & 0.667 \\
& After & \textbf{0.450} & \textbf{0.469} & \textbf{0.867} \\
\midrule
\multirow{2}{*}{ELECTRA} & Before & 0.395 & 0.406 & 0.664 \\
& After & \textbf{0.411} & \textbf{0.550} & \textbf{0.825} \\
\bottomrule
\end{tabular}
\caption{Model performance on NCTB-QA test set before and after fine-tuning}
\label{tab:main_results}
\end{table*}

BERT shows the largest gains with F1 increasing from 0.150 to 0.620, representing 313\% relative improvement. Exact Match improves from 0.059 to 0.509, an increase of 0.450 points. This dramatic change suggests that while pre-training provides weak initial transfer to Bangla educational content, BERT possesses strong adaptation capability through fine-tuning. BERTScore increases from 0.665 to 0.880, reflecting a marked enhancement in semantic answer quality.

RoBERTa demonstrates the strongest zero-shot performance, achieving an F1 score of 0.428 and an EM score of 0.427, likely due to its larger pre-training corpus and enhanced training procedures. However, gains from fine-tuning are more modest, with F1 increasing to 0.469 (9.6\% relative improvement) and EM to 0.450. Despite smaller F1 gains, RoBERTa shows substantial BERTScore improvement from 0.667 to 0.867, indicating enhanced semantic answer quality.

ELECTRA achieves an F1 of 0.550 after fine-tuning, up from 0.406, representing 35.5\% relative improvement. Exact Match improves from 0.395 to 0.411. BERTScore increases from 0.664 to 0.825. This demonstrates efficient adaptation through its replaced token detection objective.

All models show significant BERTScore improvements ranging from 0.161 to 0.215 points. Fine-tuning improves not only exact matching but also semantic answer quality. This improvement is critical for practical deployment, where paraphrased answers should be accepted, particularly important for Bangla with its rich morphological variations. 

\subsection{Performance by Question Type}
\label{subsec:by_question_type}

Table~\ref{tab:by_question_type} reports F1 scores of fine-tuned models across different question types. Overall, all models benefit from fine-tuning; however, performance varies substantially depending on the reasoning demands of each question category.

\begin{table*}[t]
\centering
\small
\begin{tabular}{lccccccc}
\toprule
\textbf{Model} & \textbf{Stage} & \textbf{Factual} & \textbf{Confirm.} & \textbf{Causal} & \textbf{Compar.} & \textbf{Negation} & \textbf{Others} \\
\midrule
BERT & Before & 0.150 & 0.148 & 0.156 & 0.126 & 0.159 & 0.163 \\
& After & \textbf{0.613} & \textbf{0.679} & \textbf{0.517} & \textbf{0.767} & \textbf{0.673} & \textbf{0.610} \\
\midrule
RoBERTa & Before & 0.396 & 0.437 & 0.440 & 0.672 & 0.482 & 0.380 \\
& After & \textbf{0.443} & \textbf{0.500} & \textbf{0.444} & \textbf{0.688} & \textbf{0.497} & \textbf{0.438} \\
\midrule
ELECTRA & Before & 0.379 & 0.410 & 0.424 & 0.617 & 0.449 & 0.355 \\
& After & \textbf{0.551} & \textbf{0.575} & \textbf{0.454} & \textbf{0.701} & \textbf{0.601} & \textbf{0.501} \\
\bottomrule
\end{tabular}
\caption{F1 scores across question types before and after fine-tuning. All models show consistent gains, with BERT achieving the largest improvements.}
\label{tab:by_question_type}
\end{table*}

BERT demonstrates the most substantial improvements across all question types, with gains ranging from +0.361 (Causal) to +0.641 (Comparative). The model achieves 0.767 F1 on Comparative questions after fine-tuning. Fine-tuning effectively addresses BERT's initial weakness in comparative reasoning.

RoBERTa, despite starting from a stronger baseline, shows more modest improvements. The model exhibits minimal gains on Causal (+0.004) and Comparative (+0.016) questions. This indicates that its pre-training already equipped it with robust capabilities for these reasoning tasks. Improvements are more substantial on Confirmation (+0.063) and Others (+0.058) categories.

ELECTRA occupies a middle ground, with improvements ranging from +0.030 (Causal) to +0.172 (Factual). The model shows particularly strong gains on Factual, Confirmation, and Negation questions, with limited improvement on Causal questions, similar to RoBERTa.

\section{Discussion}

The performance gap between zero-shot and fine-tuned models reveals a fundamental challenge in adapting English-pretrained models to typologically distinct languages. While models like RoBERTa demonstrate reasonable zero-shot capabilities, the Bangla-specific linguistic phenomena (SOV word order, extensive agglutination, and case markers) require substantial task-specific adaptation. The divergent learning curves across models suggest that pre-training objectives matter a lot. BERT's masked language modeling shows greater plasticity for downstream adaptation. Again, RoBERTa's and ELECTRA's discriminative objectives provide stronger initial representations but limited headroom for improvement.

The near-identical performance of all models on unanswerable questions (0.82--0.98 F1) contrasts sharply with their variable performance on answerable questions. Abstention is a simpler learned behavior than accurate extraction. This asymmetry has practical implications. Deploying these models in educational settings would yield high precision on unanswerable detection but inconsistent answer quality on answerable questions.

Our dataset includes chain-of-thought (CoT) reasoning annotations that remain unexplored in this work. Future research should leverage these CoT annotations with generative models such as Qwen 2.5-7B, training on (context, question) inputs with (CoT, answer) labels to enable interpretable reasoning processes \citep{wei2023chainofthoughtpromptingelicitsreasoning}. Additionally, monolingual Bangla pre-training or adapter-based methods could close the remaining performance gap. The dataset's educational domain also enables exploration of pedagogical applications, such as automated assessment generation or personalized learning systems tailored to Bangladesh's national curriculum. Finally, as the current release is purely text-based, extending NCTB-QA into a multimodal benchmark incorporating diagrams, tables, and textbook figures represents an important direction for future work.

\section{Conclusion}

NCTB-QA addresses a critical gap in Bangla NLP resources by providing a large-scale, educationally grounded question answering benchmark with balanced answerability distribution and chain-of-thought reasoning annotations. Our evaluation establishes baseline performance with extractive models and demonstrates that substantial improvements are achievable through fine-tuning. However, there remains significant room for advancing Bangla reading comprehension. We release NCTB-QA publicly to accelerate research in low-resource language understanding and educational technology for Bangla-speaking communities.

\section*{Dataset}
The dataset used in this study is not publicly available as it is currently being utilized in ongoing research projects.

\section*{Limitations}

The maximum answer length of 30 tokens, constrained by 12GB VRAM, may exclude longer educational explanations. Our evaluation focuses exclusively on encoder-only extractive models and does not utilize the chain-of-thought(CoT) annotations present in the dataset. Future work with generative models could leverage these CoT reasoning paths. Single-domain data (NCTB textbooks) may limit cross-domain generalization. LLM-based question generation could introduce systematic artifacts. Extractive QA scope excludes multi-hop and abstractive reasoning tasks.

\section*{Ethics Statement}

NCTB-QA uses publicly available educational materials with no personal data. Intended for research advancing Bangla NLP and educational technology. Users must validate model outputs before educational deployment, as incorrect answers could mislead learners.

\section*{Acknowledgments}

We thank the National Curriculum and Textbook Board of Bangladesh for open educational resources.

\bibliographystyle{plainnat}
\bibliography{custom}

\end{document}